\documentclass[10pt,twocolumn,letterpaper]{article}

\usepackage{cvpr}              

\definecolor{cvprblue}{rgb}{0.21,0.49,0.74}
\usepackage[pagebackref,breaklinks,colorlinks,allcolors=cvprblue]{hyperref}
\usepackage{pifont}
\usepackage{multirow}
\usepackage{amsmath}
\usepackage{algorithm}
\usepackage{algpseudocode}
\usepackage{amssymb}    
\usepackage{graphicx}    
\usepackage{booktabs}    
\usepackage[capitalize,noabbrev]{cleveref}
\crefname{figure}{Fig.}{Figs.}
\Crefname{figure}{Fig.}{Figs.}
\crefname{table}{Tab.}{Tabs.}
\Crefname{table}{Tab.}{Tabs.}
\crefname{equation}{Eq.}{Eqs.}
\Crefname{equation}{Eq.}{Eqs.}
\crefname{section}{Sec.}{Secs.}
\Crefname{section}{Sec.}{Secs.}
\crefname{subsection}{Sec.}{Secs.}
\Crefname{subsection}{Sec.}{Secs.}
\def\paperID{2338} 
\def\confName{CVPR}
\def\confYear{2026}

\title{Event-Illumination Collaborative Low-light Image Enhancement with a High-resolution Real-world Dataset}

\author{Senyan Xu$^{*}$, Zhijing Sun\footnotemark[1], Kean Liu, Xin Lu, Ruixuan Jiang, Mingyang Huang, \\ Xueyang Fu, Zheng-Jun Zha\footnotemark[2]\\
University of Science and Technology of China\\
{\tt\small \{syxu,sunzhijing\}@mail.ustc.edu.cn, \{xyfu,zhazj\}@ustc.edu.cn}
}

\begin{document}
\maketitle
\footnotetext[1]{Equal contribution.}
\footnotetext[2]{Corresponding author.}

\begin{abstract}
Event-based low-light image enhancement (LIE) methods mainly focus on incorporating high dynamic range (HDR) information from events while overlooking the essential global illumination in images and the inherent noise sensitivity of event signals in real-world scenarios. 
To address these issues, we propose EIC-LIE, an event-illumination collaborative LIE framework. Concretely, we first design an Event-Illumination Collaborative Interaction (EICI) module, which contains two key processes: forward gathering, which gathers HDR features across varying lighting conditions, and backward injection, which provides complementary content for illumination and event representations.
Next, we introduce an Illumination-aware Event Filter (IAEF) that dynamically reduces event noise based on brightness statistics derived from images. 
Additionally, we build a beam-splitter-based hybrid imaging system to collect high-quality event-image pairs with temporal synchronization from dynamic scenes, providing the first high-resolution, real-world event-based LIE dataset.
Extensive experiments show that our EIC-LIE outperforms state-of-the-art methods on five real-world and synthetic datasets, significantly surpassing previous methods with improvements of up to 1.24dB in PSNR and 0.069 in SSIM. The code and dataset are released at https://github.com/QUEAHREN/EIC-LIE.

\end{abstract}

\section{Introduction}

\label{sec:intro}

In low-light environments, captured images often suffer from poor visibility, increased noise, and texture loss due to the limitations of traditional sensors. These issues not only impair human visual perception but also negatively impact high-level vision tasks such as semantic segmentation~\cite{zheng2021setr}, object detection~\cite{redmon2016yolo,wang2025unleashing}, and tracking~\cite{wojke2017deepsort}. The rapid development of deep learning has significantly enhanced traditional image enhancement techniques~\cite{Wang2022Uformer, zamir2022restormer, tu2022maxim, guo2025mambair}. Specifically, recent low-light image enhancement (LIE) methods~\cite{Cai_2023_Retinexformer, weng2024mamballie} have achieved promising results by leveraging Retinex theory~\cite{land1971retinex}. These methods incorporate illumination priors to guide the enhancement network, thereby improving the overall quality of the images.

\begin{figure}
  \begin{center}
		\begin{tabular}[t]{c} \hspace{-5.8mm} 
			\includegraphics[width=0.50\textwidth]{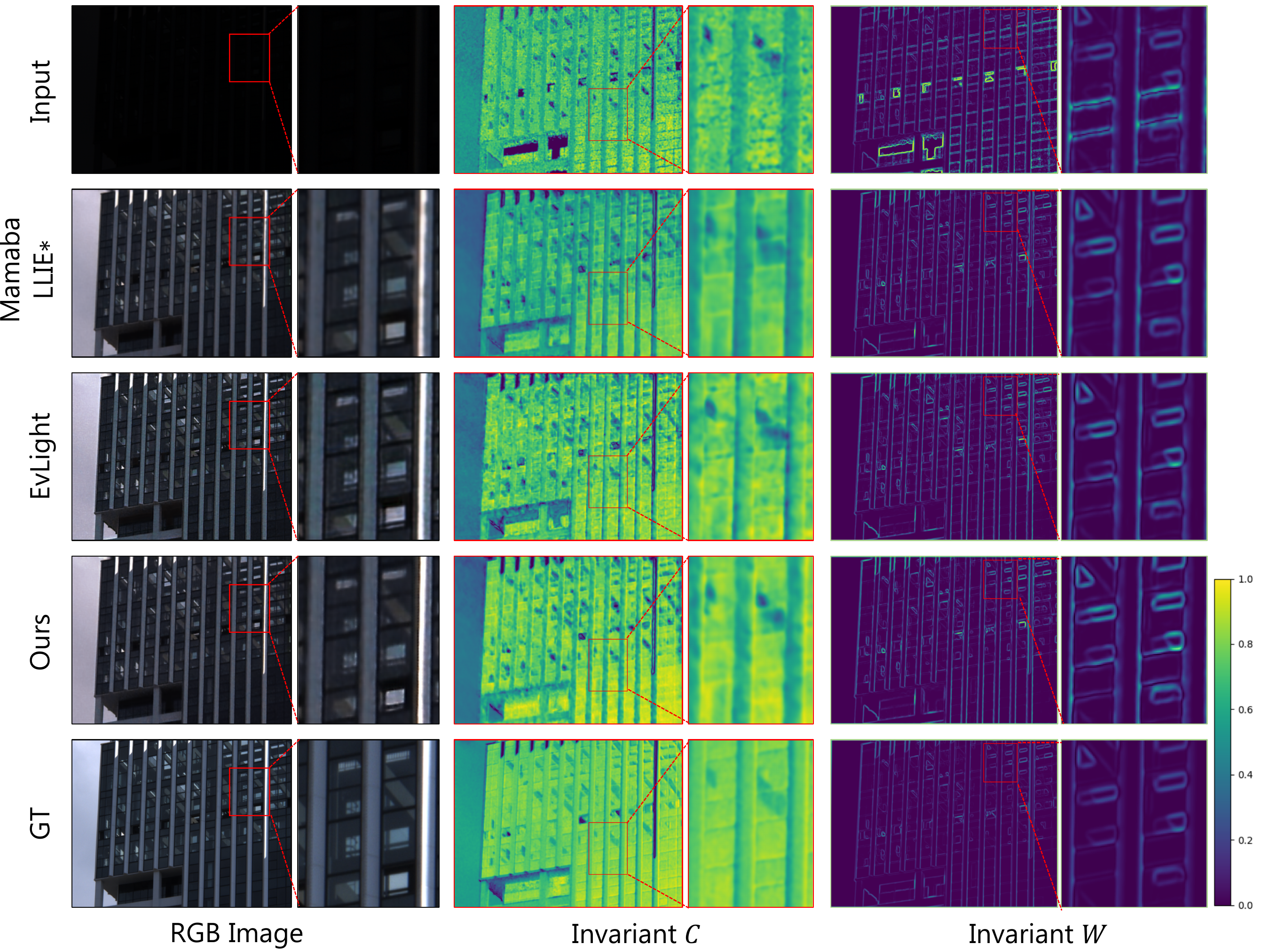}
		\end{tabular}
	\end{center}
	\vspace{-6mm}
    \caption{Visual comparison of LIE methods on the proposed real-world RLE dataset. To better illustrate the enhancement effects, color invariants \cite{geusebroek2001color} are adopted as visualization tools. Specifically, the invariant $C$ can be interpreted as describing object color regardless of intensity, while $W$ functions as an edge detector specific to changes in spectral distribution. See \emph{supp.} for more details regarding the invariants.}
	\label{fig:intro}
	\vspace{-1.6em}
\end{figure}

Compared to traditional frame-based cameras, event cameras~\cite{zhang2020learning,liu2024seeing} offer unique advantages in low-light conditions by capturing high dynamic range (HDR), high temporal resolution, and low-latency event streams. 
Nevertheless, current event-based LIE approaches still encounter multiple challenges: (1) Existing event-based LIE methods ~\cite{Liu_2023_aaai_eift, Jiang_2024_ELIE, Liang_2024_EvLight} heavily rely on direct feature fusion strategies, using complex fusion modules to compensate for missing image details. However, these methods often  \textbf{overlook the essential global illumination information emphasized in Retinex-based LIE techniques.} (2) Event-based LIE is \textbf{constrained by the inherent noise sensitivity of event signals} in low-light conditions.

As the number of available photons decreases, especially with low event-triggering thresholds ($c$ in Eq. \ref{eq:event_trigger})—
random noise significantly increases~\cite{duan2024led}. \citet{Liang_2024_EvLight} propose selectively fusing events from high Signal-to-Noise Ratio (SNR) regions, using an SNR map estimated from images. However, this fixed-guidance strategy lacks reliability, leading to noticeable noise in the enhanced results, as demonstrated in \cref{fig:intro}. 
(3) \textbf{The lack of high-resolution, high-quality, real-world datasets} has significantly hindered the progress of event-based LIE (see \cref{tab:ade}). While SDE dataset \cite{Liang_2024_EvLight} leverages a robotic arm to capture ground truth (GT) images along the same paths, their method is restricted to static scenes with slow camera movements, unavoidable temporal misalignment ($<10ms$), and severe color inaccuracies and low resolution due to the DAVIS346 sensor~\cite{Davis346}.

To address these challenges, we propose EIC-LIE, an Event-Illumination Collaborative Low-light Image Enhancement (LIE) framework. It consists of two key components, including \textbf{Event-Illumination Collaborative Interaction (EICI)} and \textbf{Illumination-Aware Event Filter (IAEF) }. The former enables collaborative bidirectional interaction between event and illumination information, merging high dynamic range (HDR) details with global illumination priors. The later dynamically filters event noise by utilizing bright statistics derived from RGB images. 
The visual analysis by introducing several physical invariants \cite{geusebroek2001color} has been shown in \cref{fig:intro}, EvLight \cite{Liang_2024_EvLight} exhibits severe noise interference in invariant $C$, while MambaLLIE \cite{weng2024mamballie} lacks HDR detail replenishment and presents relatively smooth edges in invariant $W$. Through the iterative interaction of the proposed two modules, our framework achieves accurate texture details and effective noise suppression, leading to enhanced image quality.

\begin{table}
    \centering
    \vspace{-0.3cm}
    \renewcommand\arraystretch{1.1}
\setlength{\tabcolsep}{2.5pt}
    \scalebox{0.8}{
    \hspace{-0.4cm}
    \begin{tabular}{l|cc|cc|cc}
    \hline
    \multirow{2}{*}{Dataset}  & \multirow{2}{*}{Release}   & \multirow{2}{*}{GT} &  \multicolumn{2}{c|}{Resolution} & \multicolumn{2}{c}{Dynamic}      \\ 
    ~&~&~&Event&Frame&Scene&Camera \\ \hline
    \citet{zhang2020learning}  & \ding{55} &\ding{55} & 180$\times$240 & 180$\times$240  & \ding{51} & \ding{51}   \\
     LIE~\cite{Jiang_2024_ELIE}   & \ding{51}   & \ding{51}   & 346$\times$260   & 346$\times$260 & \ding{55} & \ding{55}             \\
      \citet{liang2023coherent}             & \ding{55}   & \ding{55}        & 346$\times$260 & 1920$\times$1280 & --- & ---           \\
     SDE~\cite{Liang_2024_EvLight}   & \ding{51}   & \ding{51}   & 346$\times$260  & 346$\times$260 & \ding{55} & \ding{51}           \\ \hline
      \textbf{RLE (Ours) }  & \ding{51}  & \ding{51}   & 1024$\times$768 & 1024$\times$768 & \ding{51} & \ding{51}             \\\hline
    \end{tabular}
    }
    \vspace{-0.3cm}
    \caption{\small Summary of existing real-world event-based low-light enhancement datasets (\cref{sup:dataset_related}).}
    \vspace{-0.6cm}
    \label{tab:ade}
\end{table}

In addition, we design an optical imaging system with dual beam splitters and construct a high-resolution ($1024\times768$) RLE dataset, significantly surpassing the SDE dataset ($346\times260$) in resolution. This dataset includes high-quality event-image pairs spanning a wide illumination range in both indoor and outdoor environments, as well as complex dynamic scenes. It also features time-synchronized low-light images and event streams alongside corresponding normal-light images. Our comprehensive dataset overcomes the limitations in this field~\cite{zhang2020learning, liang2023coherent, Jiang_2024_ELIE, Liang_2024_EvLight}, providing a robust benchmark for event-based low-light enhancement. Overall, our contributions are summarized as follows:
\begin{itemize}
    \item We propose a new event-illumination collaborative low-light image enhancement (EIC-LIE) framework.
    \item We formulate collaborative bidirectional interaction processes and design Event-Illumination Collaborative Interaction (EICI) that merges high dynamic range (HDR) features across various lighting conditions and complements modal-specific content for illumination and event representations.
    \item We develop the Illumination-Aware Event Filter (IAEF) to dynamically filter event noise using bright statistics from images.
    \item We construct a high-resolution real-world dataset (RLE) consisting of high-quality image-event pairs for evaluating the proposed framework.
\end{itemize}
Experimental results demonstrate that our EIC-LIE achieves state-of-the-art performance across various benchmarks, surpassing previous studies~\cite{Liang_2024_EvLight}. Specifically, it shows improvements of up to 0.95dB/0.0469, 1.01dB/0.0695, and 1.24dB/0.0312 in terms of PSNR/SSIM on the RLE, SDE, and SDSD datasets, respectively.

\section{Related Work}
\label{sec:formatting}

\subsection{Image-based LIE}
Traditional non-learning low-light enhancement (LIE) methods relied on hand-crafted priors \cite{arici2009histogram, nakai2013color, fxy2016weighted, guo2016lime, xu2020star}, which suffer from limited adaptability and efficiency \cite{wu2022uretinex}. With significant advancements in deep learning \cite{he2016deep,vaswani2017attention,dosovitskiy2020vit,xiao2022idt,xiao2025bayesian,peng2024efficient,peng2024lightweight,peng2024towards,peng2024unveiling,peng2025boosting,peng2025directing,peng2025pixel,pengboosting,di2025qmambabsr,jiang2024rbsformer,li2024fouriermamba,liu2025dreamuhd,lu2025evenformer}, learning-based LIE methods \cite{lore2017llnet} have shown substantial improvements, which can be further categorized into Retinex-based \cite{wei2018deep, zhang2019kindling, wu2022uretinex, Cai_2023_Retinexformer, fu2023dancing, weng2024mamballie,nie2025reparameterized} and non-Retinex-based \cite{lv2018mbllen, Wang2022Uformer, wang2023lowlight, Wu_2023_LLFlow} approaches. 
Specifically, \citet{Cai_2023_Retinexformer} propose a one-stage Retinex-based Illumination-Guided Transformer to exploit the illumination representations to direct the computation of self-attention. \citet{weng2024mamballie} incorporate illumination guidance into state space models (SSMs) to enhance the Retinex-based approach. However, the crucial illumination information has not been considered in event-based LIE.

\begin{figure*}[!t]
	\begin{center}
		\begin{tabular}[!ht]{c} \hspace{-2.8mm} 
			\includegraphics[width=0.85\textwidth]{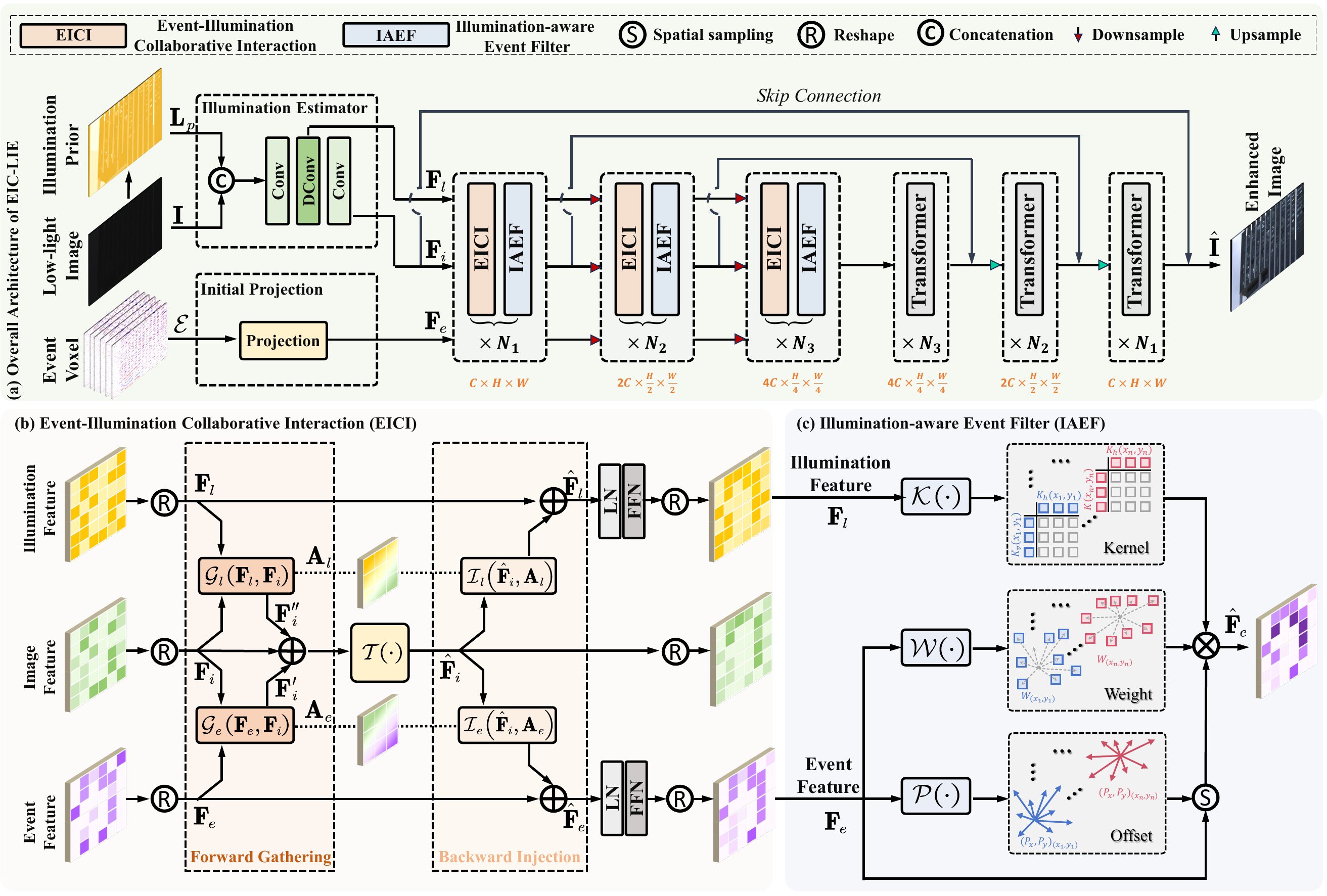}
		\end{tabular}
	\end{center}
	\vspace{-7mm}
	\caption{\small An overview of (a) our EIC-LIE. The core modules of EIC-LIE are (b) Event-Illumination Collaborative Interaction (EICI) and (c) Illumination-aware Event Filter (IAEF). Details of each module can be found in \emph{supp.}}
	\label{fig:overall}
	\vspace{-1.5em}
\end{figure*}

\subsection{Event-based LIE}
Event cameras \cite{xu2024demosaicformer,xu2025motion,liu2025event,ge2025eventmamba,cao2026learning,sun2025evdm,fu2024event,zhong2025compevent} have shown considerable potential for image enhancement \cite{wang2020eslnet, Liu_2023_aaai_eift, Jiang_2024_ELIE, Liang_2024_EvLight, weng2024evhadr_accv,sun2025low} and video enhancement \cite{liang2023coherent,zou2024eventhdr,kim2025towards}, as they offer high dynamic range (HDR) and rich edge information even in challenging low-light environments. \citet{zhang2020learning, liu2024seeing} focus on reconstructing images from low-light events but fail to recover RGB images due to the inherent lack of color information in event streams. Recent event-based LIE methods introduce RGB images and concentrate on designing complex fusion modules or strategies to achieve visually optimal results. Specifically, \citet{Jiang_2024_ELIE} design a simple residual fusion module. \citet{Liu_2023_aaai_eift} propose an Event and Image Fusion Transform module based on cross and spatial attention. \citet{liang2023coherent} first model temporal coherence by predicting motion cues from both events and frames to implement fusion,  then perform exposure correction and denoise on fused features. \citet{Liang_2024_EvLight} propose selective fusion according to the SNR map estimated by images based on Transformer.

\subsection{Event-based LIE Datasets}
\label{sup:dataset_related}
Collecting paired low-light/normal-light RGB and event data presents considerable challenges, resulting in a scarcity of real-world datasets dedicated to this purpose. \cref{tab:ade} shows the recent event-based LIE datasets. \citet{zhang2020learning} propose DVS-Dark to reconstruct intensity images from low-light events streams. \citet{Jiang_2024_ELIE} construct the LIE dataset, the first real-world dataset captured under both indoor and outdoor conditions using the DAVIS346 event camera, simulating lighting variations in static indoor and outdoor scenes by adjusting the camera's light intake and employing different exposure times, respectively. \citet{liang2023coherent} design a beam-splitter-based system to capture events and images, equipped with an industrial camera (FLIR Chameleon 3 Color) and an event camera (DAVIS346). \citet{Liang_2024_EvLight} capture SDE dataset, employing a programmable robotic arm to capture ground truth (GT) images by precisely controlling an event camera along identical trajectories. To the best of our knowledge, there is currently no dataset that simultaneously offers high-resolution and high-quality event-image pairs with GT, which significantly hinders research on event-based LIE.

\section{Methodology}

\subsection{Preliminaries}
\vspace{-1mm}
\textbf{Retinex Theory.}
Retinex theory~\cite{land1971retinex} posits that a low-light image  \(\mathbf{I} \in \mathbb{R}^{H \times W \times 3}\) can be decomposed into reflectance \(\mathbf{R} \in \mathbb{R}^{H \times W \times 3}\) and illumination maps \(\mathbf{L} \in \mathbb{R}^{H \times W \times 3}\), which can be formulated as:
\vspace{-0.75mm}
\begin{equation}
  \mathbf{I} = \mathbf{R} \odot \mathbf{L}.  
         \vspace{-0.5mm}
\end{equation}
Recent Retinex-based methods typically focus on either jointly estimating both maps \cite{wei2018deep,zhang2019kindling,wu2022uretinex} or estimating a lit-up map (with reflectance treated as the enhanced output) \cite{Cai_2023_Retinexformer} to restore normal light images \(\mathbf{N} \in \mathbb{R}^{H \times W \times 3}\). These two types of methods can be respectively formulated as:
\vspace{-0.75mm}
\begin{equation}
\mathbf{N} = \mathbf{\tilde{R}} \odot \mathbf{\tilde{L}},~~\mathbf{N} = \mathbf{I}\odot \Bar{\mathbf{L}},          \vspace{-0.5mm}
\end{equation}
where \(\odot\) denotes the element-wise multiplication; \(\Bar{\mathbf{L}}\) denotes the estimated lit-up map; \(\mathbf{\tilde{R}}\) and \(\mathbf{\tilde{L}}\) denote the estimated reflectance and illumination maps. Unlike these, \citet{weng2024mamballie} proposes a Retinex-aware selective kernel module to modulate features in the network by illumination prior implicitly.

However, these image-based Retinex-based methods are fundamentally limited by the constraints of the traditional sensor due to the illumination information of an image being restricted by the sensor's dynamic range. To address this, designing a framework that enables interaction between event and illumination information is a promising solution.

\noindent \textbf{Event Representation.}
Events are triggered when brightness changes exceed a set threshold. Formally, given an event sequence \(\mathcal{E} = \{ e_k \}_{k=1}^{N_e} \), \( e_k = \{ (x_k, y_k, t_k, p_k) \} \) is defined by pixel location \( (x_k, y_k) \), timestamp \( t_k \), and polarity \( p_k \in \{ +1, -1 \} \), which represents an increase or decrease in brightness. The $N_e$ represents the number of events. The event trigger condition can be formulated as:
\vspace{-0.75mm}
\begin{equation}
log\frac{\mathcal{L}(x_k, y_k, t_k)}{\mathcal{L}(x_k, y_k, t_k - \Delta t)} = p_k \cdot c,
\label{eq:event_trigger}
         \vspace{-0.5mm}
\end{equation}
where \( c \) is the contrast threshold, and \(\mathcal{L}\) is the brightness. The raw event stream is similar in form to point clouds but contains far more points than them. This makes it challenging to design an efficient event representation, often resulting in a time-space trade-off. In our approach, to fully leverage the high dynamic range information, we adopt the SBT (Stacking Based on Time) representation \cite{wang2019event}, stacking events into \( B \) time bins. Given event voxel $V$, the polarity accumulation for the $i$-th time bin is computed as:
\vspace{-0.75mm}
\begin{equation}
\mathcal{V}(i) = \sum_{k \in \mathcal{T}_i} p_k,
         \vspace{-0.5mm}
\end{equation}
where $\mathcal{T}_i = \left\{ k \mid t_k \in \left[ t_0 + \frac{(i-1)\Delta t}{B}, \, t_0 + \frac{i\Delta t}{B} \right) \right\}$ is the set of events within the $i$-th time interval, $\Delta t = t_{N_e} - t_0$ is the total event duration.

\subsection{Overall Pipeline}

\cref{fig:overall} illustrates the overall architecture of the proposed EIC-LIE. Our goal is to implement the interaction between illumination and events while ensuring robust performance under real-world low-light conditions. We first estimate illumination features $\mathbf{F}_l$ from the illumination prior $\mathbf{L}_p$ and extract initial event features $\mathbf{F}_e$. $\mathbf{L}_p$ is derived by applying a pixel-wise maximum operation across all channels of the image $ I  $. Then, we design two modules to achieve our goal: \textbf{(i)} Event-Illumination Collaborative Interaction (EICI), and \textbf{(ii)} Illumination-aware Event Filter (IAEF).

\subsection{Event-Illumination Collaborative Interaction}
In this section, we aim to establish collaborative bidirectional feature transmission between illumination and events. Inspired by \citet{koner2025lookupvit}, given an auxiliary feature $\mathbf{X} \in \mathbb{R}^{N \times C}$ and a primary feature  $\mathbf{T} \in \mathbb{R}^{N \times C}$, we define collaborative bidirectional interaction as a mechanism comprising two key processes:

\noindent \textbf{Forward Gathering.}  
The forward gathering process establishes a unidirectional information flow from $\mathbf{X}$ to $\mathbf{T}$ via covariance-based cross-attention mechanism \cite{ali2021xcit, zamir2022restormer}. Specifically, this process $\mathcal{G}(\cdot,\cdot)$ can be formulated as:
\vspace{-0.75mm}
\begin{equation}
    (\mathbf{T}^{\prime},\mathbf{A})  = \mathcal{G}(\mathbf{X}, \mathbf{T}),
\vspace{-0.5mm}
\end{equation}
where $\mathbf{T}' \in \mathbb{R}^{N \times C}$ represents the updated primary feature and $\mathbf{A}\in \mathbb{R}^{C \times C}$ denotes the intermediate attention matrix computed during the process. The attention-based interaction enables $\mathbf{T}$ to selectively absorb relevant contextual information from $\mathbf{X}$ while preserving spatial consistency.

\noindent \textbf{Backward Injection.}  
The primary representation is decomposed by reusing the stored attention matrix $\mathbf{A}$ to ensure modality-specific feature separation. The backward injection process $\mathcal{I}(\cdot,\cdot)$ reconstructs modality-specific components by selectively redistributing information from the primary feature to the original auxiliary feature space: 
\vspace{-0.75mm}
\begin{equation}
    \mathbf{X}^{\prime} = \mathcal{I}(\mathbf{T}^{\prime}, \mathbf{A}) + \mathbf{X},
\vspace{-0.5mm}
\end{equation}
where $\mathbf{X}^{\prime} \in \mathbb{R}^{N \times C}$ denotes the refined modality-specific features. This operation ensures that the extracted information is consistently shared while preserving modality integrity. The details of $\mathcal{G}(\cdot,\cdot)$ and $\mathcal{I}(\cdot,\cdot)$ can be found in \emph{supp.}

\begin{figure*}[!t]
	\begin{center}
		\begin{tabular}[!ht]{c} \hspace{-2.8mm} 
			\includegraphics[width=0.86\textwidth]{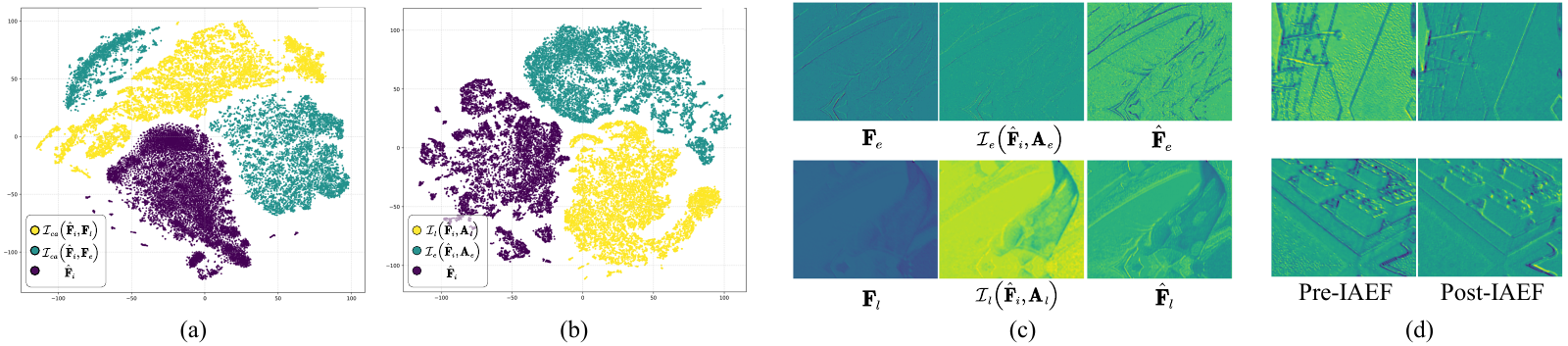}
		\end{tabular}
	\end{center}
	\vspace{-7mm} 
	\caption{\small (a) t-SNE analysis of features in EICI without attention reuse. Note that $\mathcal{I}_{ca}(\cdot,\cdot)$ here denotes the direct computation of cross-attention between the two features, corresponding to Case 8 in \cref{sec:exp_abl}.
(b) t-SNE analysis of features in EICI with attention reuse.
(c) Visualization of features in EICI.
(d) Visualization of features in IAEF, showing pre-IAEF and post-IAEF event features.
    }
	\label{fig:vis_tsne}
	\vspace{-1.5em}
\end{figure*}

\noindent\textbf{Event-Illumination Collaborative Interaction.} Based on basic collaborative bidirectional operation, we first gather reshaped illumination feature \( \mathbf{F}_{l} \in \mathbb{R}^{HW\times C} \) and event feature \( \mathbf{F_{e}} \in \mathbb{R}^{HW\times C} \) into the reshaped image feature \( \mathbf{F_{i}} \in \mathbb{R}^{HW\times C} \) to achieve event-illumination collaborative fusion.
\vspace{-0.75mm}
\begin{equation}
        (\mathbf{F}_{i}^{\prime}, \mathbf{A}_{e}) = \mathcal{G}_{e}(\mathbf{F}_{e}, \mathbf{F}_{i}),\quad(\mathbf{F}_{i}^{\prime \prime}, \mathbf{A}_{l}) = \mathcal{G}_{l}(\mathbf{F}_{l}, \mathbf{F}_{i}),
         \vspace{-0.5mm}
\end{equation}
where $\mathbf{F}_{i}^{\prime}\in \mathbb{R}^{HW\times C}$ aggregates $\mathbf{F}_{e}$ to supplement high dynamic range information, and $\mathbf{F}_{i}^{\prime \prime}\in \mathbb{R}^{HW\times C}$ further aggregates $\mathbf{F}_{l}$ to supplement global illumination information. After the gathering step, the features are refined in a latent domain to obtain a comprehensive fusion, which is formulated as:
\vspace{-0.75mm}
\begin{equation}
        \hat{\mathbf{F}}_{i} = \mathcal{T}(\mathbf{F}_{i} +\mathbf{F}_{i}^{\prime}+\mathbf{F}_{i}^{\prime\prime}),
         \vspace{-0.5mm}
\end{equation}
where \( \mathcal{T}(\cdot) \) denotes a transformer block that fuses features by self-attention in the latent space, further enhancing the interaction between illumination and event information. Then, to extract refined illumination-specific and event-specific features from the latent representation $\hat{\mathbf{F}}_{i}\in \mathbb{R}^{HW\times C}$, we employ a decomposition module to convert the latent features into specific components:
\vspace{-0.75mm}
\begin{equation}
        \hat{\mathbf{F}_{l}} = \mathcal{I}_{l}(\hat{\mathbf{F}}_{i}, \mathbf{A}_{l}) + \mathbf{F}_{l},\quad \hat{\mathbf{F}_{e}} = \mathcal{I}_{e}(\hat{\mathbf{F}}_{i}, \mathbf{A}_{e})+ \mathbf{F}_{e},
         \vspace{-0.5mm}
\end{equation}
where \( \hat{\mathbf{F}_{l}}\in \mathbb{R}^{HW\times C}  \) and \( \hat{\mathbf{F}_{e}}\in \mathbb{R}^{HW\times C} \) denote that the processed illumination and event features, respectively, obtained by the injection operations $\mathcal{I}_{l}$ and $\mathcal{I}_{e}$. Reusing the corresponding attention matrices \( \mathbf{A}_{e}\in \mathbb{R}^{C\times C} \) and \( \mathbf{A}_{l}\in \mathbb{R}^{C\times C} \) to impose implicit alignment constraints, ensures accurate extraction of the event and illumination features during the backward injection process.

\noindent \textbf{Discussion on Motivations.}  Unidirectional information injection limits true collaboration. Features from one modality are injected into the other, but lack the reciprocal exchange needed for mutual refinement. This prevents features from being dynamically updated with complementary information throughout the network. To enable event-illumination collaboration, the proposed Backward Injection facilitates bidirectional refinement by employing reused attention with implicit alignment constraints. This allows both event and illumination features to be continuously enriched by information from the other modality. \cref{fig:vis_tsne}(b) demonstrates that our method achieves more compact feature separation compared to \cref{fig:vis_tsne}(a) (which does not reuse attention, i.e., directly employs cross-attention, corresponding to Case 8 in \cref{sec:exp_abl}). \cref{fig:vis_tsne}(c) further reveals that the decomposed features $\mathcal{I}_{l}(\hat{\mathbf{F}}_{i}, \mathbf{A}_{l})$ and $\mathcal{I}_{e}(\hat{\mathbf{F}}_{i}, \mathbf{A}_{e})$   possess strong modality-specific properties while providing complementary high dynamic range textures and static brightness information.

\subsection{Illumination-aware Event Filter}

In real-world low-light scenarios, the limited number of available photons \cite{cao2023physicssensor} leads to unpredictable event spikes triggered by a small number of randomly arriving photons. This effect is particularly prominent with a low event-triggering threshold ($c$ in Eq. \ref{eq:event_trigger}), resulting in random noise within the event stream \cite{duan2024led}. Traditional event-denoising filters \cite{liu2015design,guo2022low} or framework \cite{gallego2018unifying} struggle to balance noise suppression and the preservation of meaningful events. The primary challenges lie in \textbf{(i)} the significant differences in the spatiotemporal statistical distributions of signal and noise events and \textbf{(ii)} the lack of global illumination modeling. To address these limitations, we propose an \textbf{Illumination-aware Event Filter (IAEF)} module, which introduces global illumination priors derived from images captured by a frame-based sensor. This prior provides a stable illumination statistical feature, which is further utilized to guide the dynamic weighting of the event filter. Based on this insight, inspired by the design of traditional filters\cite{rigamonti2013learningsconv,niklaus2017videoconv,liu2024motion}, our filter design incorporates the following two key components:

\begin{figure*}[!t]
	\begin{center}
		\begin{tabular}[!ht]{c} \hspace{-2.8mm} 
			\includegraphics[width=\textwidth]{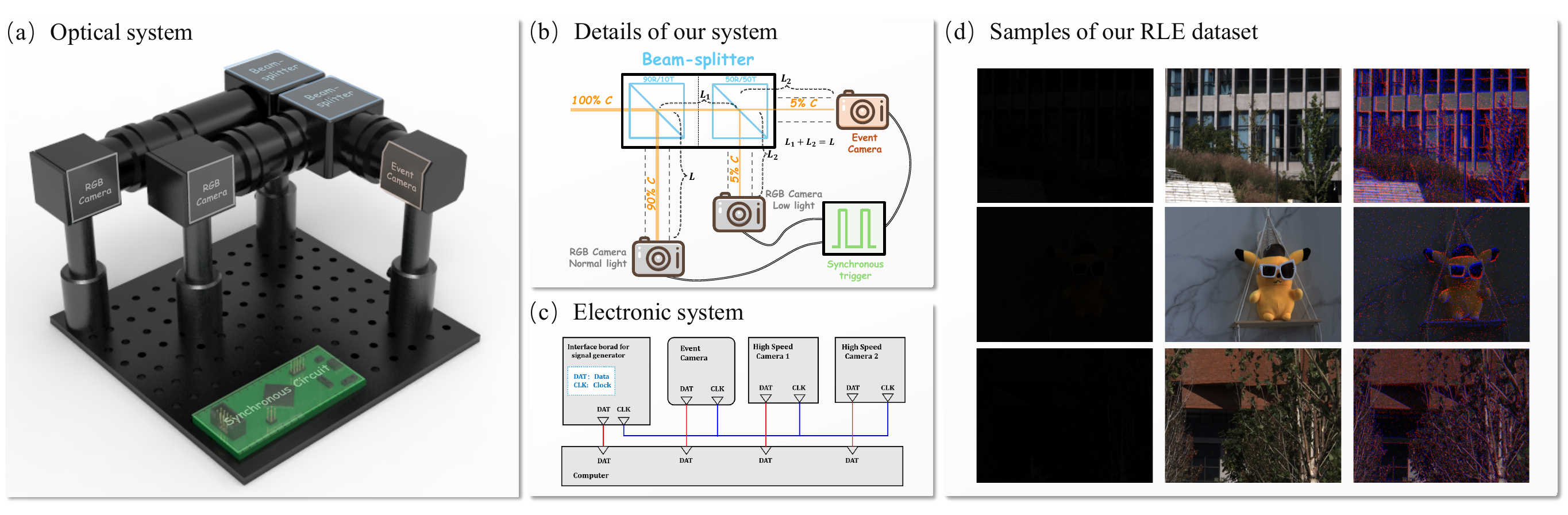}
		\end{tabular}
	\end{center}
	\vspace{-8mm} 
	\caption{\small The hardware implementation of our imaging system. In (d), from left to right, each represents low-light images, normal-light images, and aligned event streams, respectively. Refer to \emph{supp.} to find more video samples of the RLE dataset. }
	\label{fig:device}
\end{figure*}

\begin{table*}
\renewcommand\arraystretch{1.1}
\setlength{\tabcolsep}{5.5pt}
\vspace{-2mm}
    \centering

\resizebox{\textwidth}{!}{
    \begin{tabular}{l|c|c|cc|cc|cc|cc|cc}
    \toprule
         \multirow{2}{*}{\textbf{Methods}}  & \multirow{2}{*}{\textbf{Venue}} & \multirow{2}{*}{\textbf{Input}}  & \multicolumn{2}{c|}{\textbf{SDE-indoor}}  & \multicolumn{2}{c|}{\textbf{SDE-outdoor}} & \multicolumn{2}{c|}{\textbf{SDSD-indoor}} & \multicolumn{2}{c|}{\textbf{SDSD-outdoor}} & \textbf{\#Prams} & \textbf{FLOPs} \\ [0.5ex]
            ~  & ~  & ~  & \textbf{PSNR} & \textbf{SSIM} &  \textbf{PSNR} & \textbf{SSIM} & \textbf{PSNR} & \textbf{SSIM}  & \textbf{ PSNR}& \textbf{ SSIM} & \textbf{ (M)}& \textbf{(G)}\\
         \hline        
          E2VID+\cite{stoffregen2020reducing} &ECCV'20  &E &15.19  &0.5891  &15.01  &0.5765 &13.48 &0.6494 &16.58 &0.6036 &10.71 & 27.99    \\
          \hline
          SNR-Net\cite{Xu_SNRNet} &CVPR'22 & I  &20.05  &0.6302  &22.18  &0.6611    &24.74 &0.8301 &24.82 &0.7401&4.01 & 26.35 \\
          Uformer\cite{Wang2022Uformer} &CVPR'22  & I &21.09  &0.7524  &22.32    & 0.7469   &24.03 &0.8999 &24.08 &0.8184&5.29 & 12.00 \\
          
         LLFlow-L-SKF\cite{Wu_2023_LLFlow}  & CVPR'23  &I & 20.92  &0.6610  &21.68  & 0.6467 &23.39 &0.8180 &20.39 &0.6338&39.91 & 409.50 \\
         Retinexformer\cite{Cai_2023_Retinexformer} &ICCV'23  &I  &21.30  &0.6920 &22.92  &0.6834   &25.90 &0.8515 &26.08 &0.8150&1.61 & 15.57 \\
         SFHFormer\cite{jiang2024sfhformer} &ECCV'24  &I  &20.98 & 0.6775 &23.01  &0.7534 & 26.39 & 0.8956& 23.26& 0.7539 &1.54 & 19.61 \\
         MambaLLIE\cite{weng2024mamballie} &NIPS'24  &I & 21.37  &0.7050 &21.86  &\underline{0.7591}  &27.76 &0.9042& 25.50  &0.8023 &2.28 & 20.85 \\
         \hline
         ELIE\cite{Jiang_2024_ELIE} & TMM'23  &I+E &19.98  &0.6168  &20.69  & 0.6533  &27.46 &0.8793 &23.29 &0.7423&33.36 & 440.32 \\
         eSL-Net\cite{wang2020eslnet} & ECCV'20  &I+E &21.25  &0.7277  &22.42  & 0.7187&24.99 &0.8786 &24.49 &0.8031&0.56 & 560.94 \\
         Liu et al.\cite{Liu_2023_aaai_eift} & AAAI'23  &I+E &21.79  &0.7051  &22.35  & 0.6895   &27.58 &0.8879 &23.51 &0.7263 &47.06 & 44.71 \\ 
         EvLowight\cite{liang2023coherent} & ICCV'23  &I+E*  &20.57  &0.6217  &{22.04}& 0.6485   &{23.14} &{0.8143} &{23.27} &{0.7363}&15.03 & - \\ 
         EvLight\cite{Liang_2024_EvLight} & CVPR'24  &I+E  &\underline{22.44}  &\textbf{0.7697}  &\underline{23.21}& 0.7505   &\underline{28.52} &\underline{0.9125} &\underline{26.67} &\underline{0.8356}&22.73 & 180.90 \\    
         \hline
         \textbf{Ours} & -  &I+E &\textbf{23.33} & \underline{0.7573}  &\textbf{24.22}  & \textbf{0.8200} &\textbf{29.76}  & \textbf{0.9193}   &\textbf{27.45}   &  \textbf{0.8668}  &2.13 & 70.95 \\
    \bottomrule
    \end{tabular}
    }
        \caption{\small The quantitative results on SDE-indoor, SDE-outdoor, SDSD-indoor, and SDSD-outdoor test datasets. Note that 'E', 'I', and 'I+E' represent the input type corresponding to event-only, image-only, and event-image, respectively. FLOPs are estimated with the resolution of $256\times256$. The best and
the second results are boldfaced and underlined, respectively.  }
    \label{tab_com_all}
    \vspace{-1em}
\end{table*}

\noindent \textbf{Illumination-aware Kernel Extraction (\( k \)).} 
Given the illumination feature $\mathbf{F}_l \in \mathbb{R}^{C\times H\times W}$, we calculate the $n\times n$ kernel \(\mathbf{K}\) by:
\vspace{-0.75mm}
\begin{equation}
    \mathbf{K}_v,\mathbf{K}_h = \mathcal{K}(\mathbf{F}_l),
\end{equation}
where $\mathcal{K(\cdot)}$ is the kernel extraction module, $\mathbf{K}_v\in \mathbb{R}^{n\times H\times W}$ and $\mathbf{K}_h\in \mathbb{R}^{n\times H\times W}$ are the 1D filter kernels in the vertical and horizontal directions corresponding to $\mathbf{K}\in \mathbb{R}^{n^2\times H\times W}$. By incorporating the illumination-aware distribution into the kernel, \( \mathbf{K} \) encodes additional lighting priors for subsequent filtering. This illumination-aware representation enhances the filter's robustness, enabling it to distinguish between noise and meaningful event responses.

\noindent\textbf{Event-driven Weight and Offset Extraction (\( \mathbf{W} \) and \( (\mathbf{P}_x, \mathbf{P}_y) \)).}  Given the event feature $\mathbf{F}_e \in \mathbb{R}^{C\times H\times W}$, we calculate the $n\times n$ weight \(w\) and x/y-axis offset \( (\mathbf{P}_x, \mathbf{P}_y) \) by:
\vspace{-0.75mm}
\begin{equation}
    \mathbf{W} = \mathcal{W}(\mathbf{F}_e), \quad \mathbf{P}_x,\mathbf{P}_y = \mathcal{P}(\mathbf{F}_e),
\end{equation}
where $\mathcal{W(\cdot)}$ and $\mathcal{P(\cdot)}$ are the weight extraction and offset prediction modules, respectively. $\mathbf{W}\in \mathbb{R}^{n^2\times H\times W}$ is utilized to modulate the contribution of different referenced events adaptively, quantifying the reliability of events occurring at a given pixel or within its local neighborhood. Meanwhile, $\mathbf{P}_x\in \mathbb{R}^{n^2\times H\times W}$ and $\mathbf{P}_y\in \mathbb{R}^{n^2\times H\times W}$ define the spatial offsets, dynamically referencing the coordinates of available neighboring events, enabling the filter to flexibly select informative events, thereby improving robustness against noise and misaligned event locations.

Based on these components, given the pixel coordinate $(m,n)$ in $\mathbf{F}_e$ , our IAEF can be formulated as:
\vspace{-0.5mm}
\begin{equation}
\begin{aligned}
    \hat{\mathbf{F}}_e(m, n) = \sum_{(\mathbf{P}_{x}(m,n),\mathbf{P}_{y}(m,n))} \mathbf{W}(m,n) \cdot \mathbf{K}(m,n) \\
    \cdot \mathcal{S}\Bigl( \mathbf{F}_e, \bigl(m + \mathbf{P}_{x}(m,n), n + \mathbf{P}_{y}(m,n)\bigr) \Bigr),
\end{aligned}
\end{equation}
where $\mathcal{S(\cdot,\cdot)}$ denotes the spatial sampling operation, $\hat{\mathbf{F}}_e(m, n)$ denotes the output feature, $\mathbf{K}(m,n)$ is approximated by $\mathbf{K}_v(n)\cdot \mathbf{K}_h(m)$. As shown in \cref{fig:vis_tsne}(d), the event feature processed by IAEF (Post-IAEF) exhibits lower noise levels compared to that without IAEF (Pre-IAEF).


\section{Our RLE Dataset}

Compared to methods that directly reconstruct intensity frames from event streams \cite{stoffregen2020reducing, liu2024seeing}, incorporating RGB image enables low-light enhancement for color images. However, acquiring paired dynamic sequences in real-world conditions remains a significant challenge. Early traditional frame-based methods captured paired data using stereo systems \cite{Ignatov_2017_DSLR}, systems with beam-splitters \cite{jiang2019learningtosee,lee2023human}, and repeatable electromechanical systems \cite{wang2021sdsd}. Recently, \citet{Liang_2024_EvLight} has provided a large-scale real-world event-image dataset (SDE Dataset) by designing a robotic alignment system equipped with the DAVIS346 event camera, capable of simultaneously capturing events and frames under low-light and normal-light conditions. Nonetheless, it still faces inherent issues: i) the camera controlled by the robotic arm only captures the data with camera movements, prohibiting other relative motions in the scene; ii) there are inevitable temporal errors in sequences captured multiple times; iii) although the DAVIS346 can record both events and frames, its output frames exhibit low resolution and color distortions \cite{Davis346}, as shown in \emph{supp}. These limitations significantly hinder its applicability in real-world scenarios.

\begin{figure*}[t]

	\begin{center}
		\begin{tabular}[t]{c} \hspace{-4mm}
        
			\includegraphics[width=0.78\textwidth]{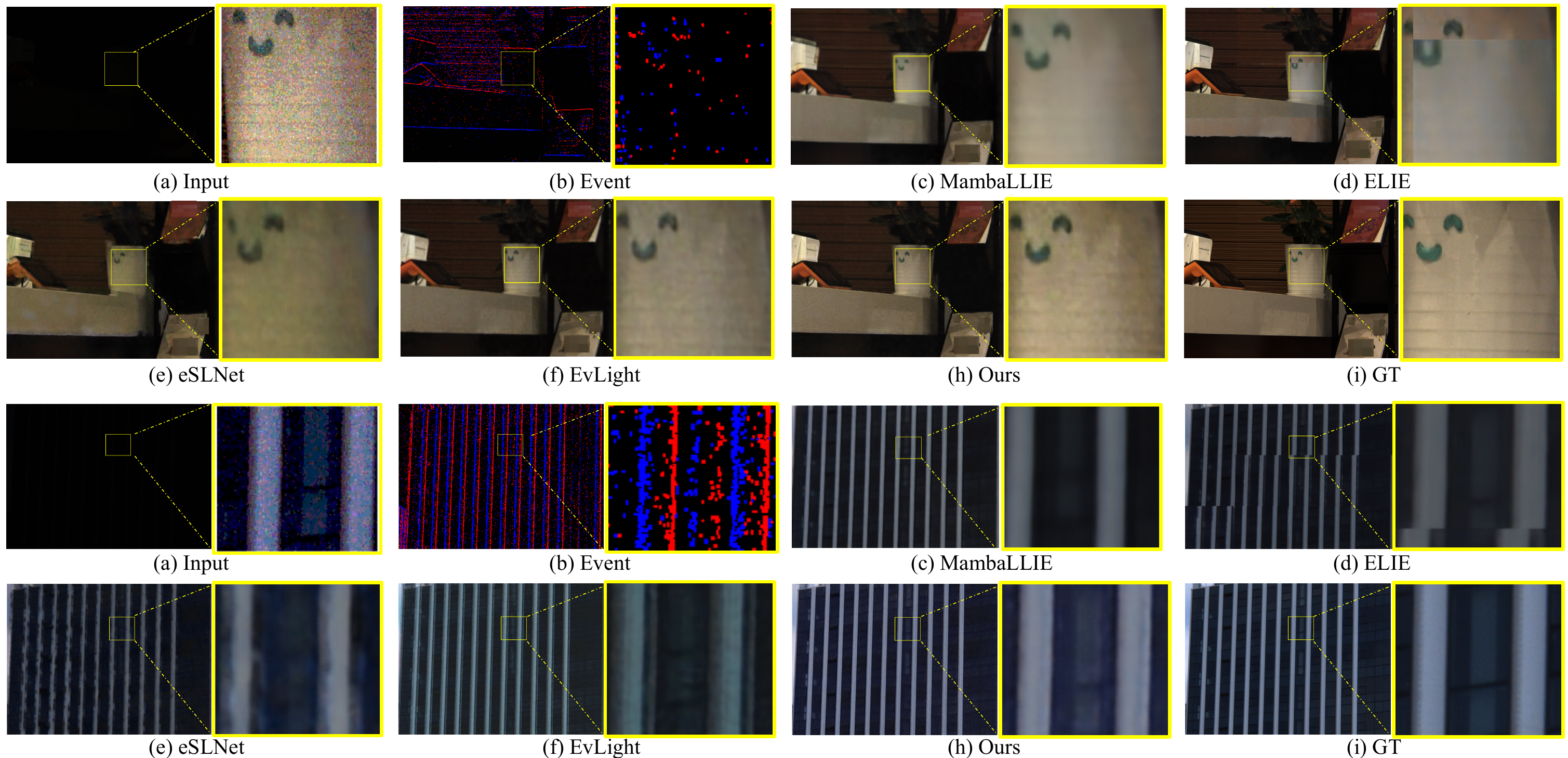}
		\end{tabular}
	\end{center}
\vspace*{-6mm}
\caption{\small Visual results on RLE dataset. Note that the crop of input has been gamma corrected, and other figures also follow this adjustment. Zoom in for a better view.}
	\label{fig:compare_rle}
	\vspace{-1.0em}
\end{figure*}

To address these problems, we design an optical system based on dual beam splitters to achieve coaxial alignment of multiple cameras, as illustrated in \cref{fig:device}(a). This system is equipped with two high-resolution RGB cameras (featuring the Sony IMX273 sensor, with an output resolution of $1440\times 1080$) and an advanced event camera (featuring the Sony\&Prophesee IMX636 sensor, with an output resolution of $1280\times 720$). The first beam-splitter has a specification of 10R/90T, where 90\% of the light is captured by the RGB camera A for normal light conditions. The remaining 10\% of the light enters the second 50R/50T beam-splitter, where the irradiance for the low-light RGB camera and the event camera is attenuated to $10\% \times 50\%$. 
As shown in \cref{fig:device}(b), we equipped the normal-light camera with an optical sleeve of a specific length to align the optical path configuration, enhancing spatial alignment accuracy. Building upon this precisely designed physical alignment, we further employed the commonly used homography transformation to ensure spatial consistency between events and images. To achieve temporal alignment of the captured sequences, an external synchronization controller is utilized to ensure precisely, synchronized triggering of the three cameras, as present in \cref{fig:device}(c). Our imaging system enables the simultaneous capture of high-quality low-light images, event streams, and paired normal-light images, even in complex dynamic scenes commonly encountered in real-world environments. In \cref{fig:device}(d), we present three examples from our dataset. More details about RLE can be found in \emph{supp.}

\begin{table}
    \centering
    \renewcommand\arraystretch{1.1}
    \scalebox{0.75}{
    \hspace{-0.4cm}
  \begin{tabular}{l|c|cc|c}
    \toprule
         \multirow{2}{*}{\textbf{Methods}}  & \multirow{2}{*}{\textbf{Input}} & \multicolumn{2}{c|}{\textbf{RLE}}& \textbf{Runtime}      \\[0.5ex]
          ~  & ~  & \textbf{PSNR} & \textbf{SSIM}  & \textbf{(ms)} \\
         \hline        
   Retinexformer\cite{Cai_2023_Retinexformer}  &I  &20.33 & 0.6858 &105 \\
         SFHFormer\cite{jiang2024sfhformer}   &I  &19.80 & 0.6827 &108 \\
         MambaLLIE\cite{weng2024mamballie}  &I  &21.40 & 0.7389 &386   \\
         \hline
         ELIE\cite{Jiang_2024_ELIE}    &I+E & 20.88 & \underline{0.7655}&936  \\
         eSL-Net\cite{wang2020eslnet}   &I+E& 19.32& 0.6805 &191 \\ 
         EvLowlight \cite{liang2023coherent}  &I+E* & 19.10& 0.7122&- \\    
         EvLight \cite{Liang_2024_EvLight}  &I+E & \underline{22.68}& 0.7201&323 \\    
         \hline
         \textbf{Ours}  &I+E & \textbf{23.63}&\textbf{0.7670} &298   \\
    \bottomrule
    \end{tabular}
    }
    \caption{\small The quantitative results on RLE test datasets. The average runtime is computed for an image size of $1024\times768$, on an NVIDIA 4090 GPU. EvLowlight \cite{liang2023coherent} is a Video-based$^*$ method.} 
            \label{tab_com_RLE}
                \vspace{-1.5em}
\end{table}

\section{Experiments} 
\vspace{-1mm}

\subsection{Comparison}

\begin{figure*}[t]
	\begin{center}
		\begin{tabular}[t]{c} \hspace{-4mm}
			\includegraphics[width=\textwidth]{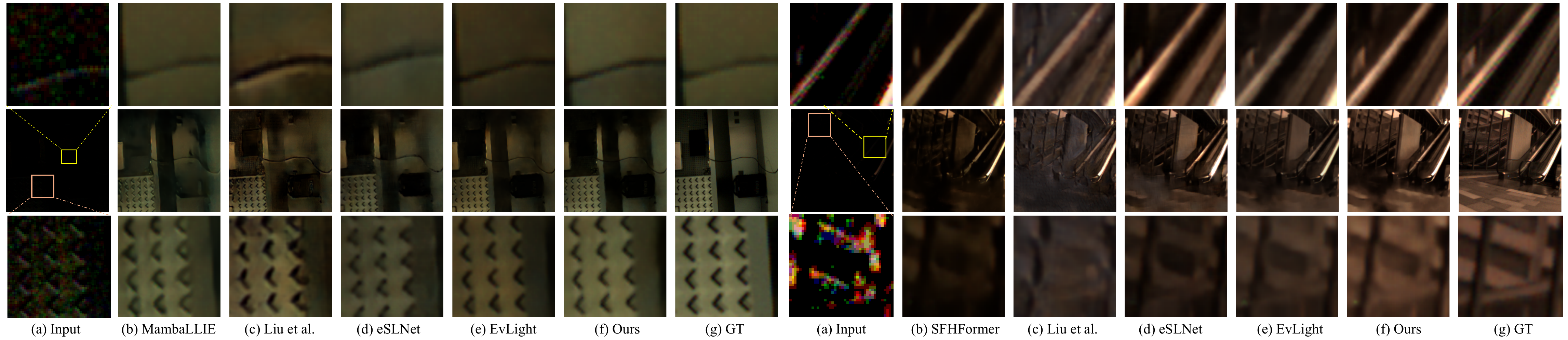}
		\end{tabular}
	\end{center}
	\vspace*{-6mm}
	\caption{\small Visual results on SDE~\cite{Liang_2024_EvLight}-indoor (left) and -outdoor (right). Zoom in for a better view.}
	\label{fig:compare_sde}
\end{figure*}

\begin{figure*}[t]
	\begin{center}
		\begin{tabular}[t]{c} \hspace{-4mm}
			\includegraphics[width=\textwidth]{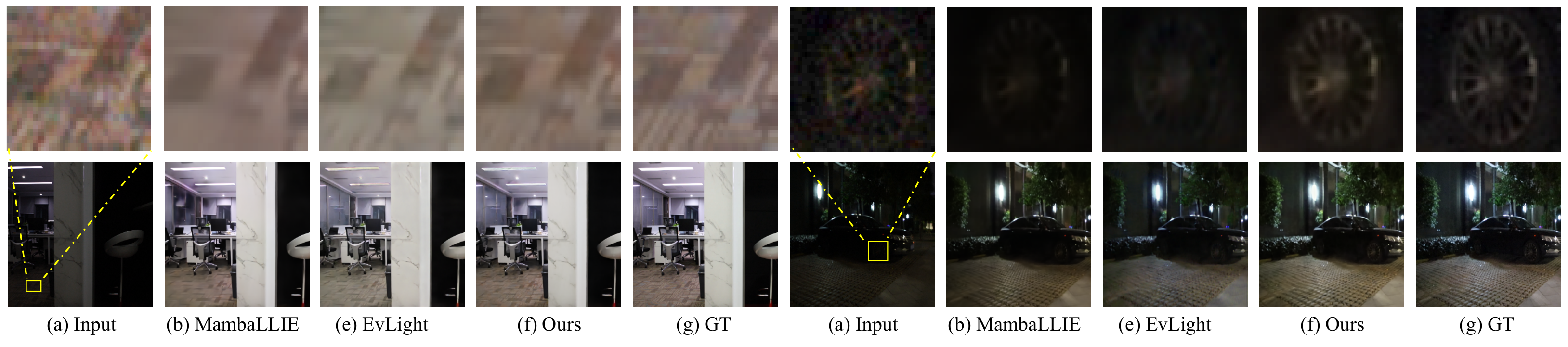}
		\end{tabular}
	\end{center}
	\vspace*{-6mm}
	\caption{\small Visual results on SDSD~\cite{wang2021sdsd}-indoor (left) and -outdoor (right). Zoom in for a better view.}
	\label{fig:compare_sdsd}
\end{figure*}

\begin{table*}[pbt!]
\parbox{.32\textwidth}{
\centering

\scalebox{0.9}{
\begin{tabular}{c|c|cc}
                \toprule
                \multirow{2}{*}{\textbf{Case}}& \multirow{2}{*}{\textbf{Guidance}}& \multirow{2}{*}{\textbf{PSNR}} & \multirow{2}{*}{\textbf{SSIM}} \\[0.5ex]
                ~  & ~  & ~ &~  \\
                \hline        
                0  & - & 19.53& 0.6623 \\
                
                1 & w. E & 21.40& 0.7287\\
                2  & w. I & 21.08& 0.7194 \\
                \hline
                \textbf{Ours}  & w. I\&E & \textbf{23.63} & \textbf{0.7670}  \\
                \bottomrule
            \end{tabular}}
            \vspace{-2.2mm}
            \captionof{table}{\small Ablation study for the event and illumination guidance. Case 0 is the baseline.} 
\label{tab_abl1}

}
\hfill
\parbox{.31\textwidth}{
\centering

\scalebox{0.9}{
\begin{tabular}{c|c|cc}
                \toprule
                \multirow{2}{*}{\textbf{Case}}& \textbf{Event} & \multirow{2}{*}{\textbf{PSNR}} & \multirow{2}{*}{\textbf{SSIM}} \\[0.5ex]
                ~  & \textbf{Filter}   & ~&~  \\
                \hline
                  3  & - & 20.92 & 0.7064 \\
                  4  &w. Conv. & 21.92& 0.7481 \\
                  5  &w. TB & 21.86& 0.7430 \\

                \hline
                \textbf{Ours}  &w. IAEF & \textbf{23.63} & \textbf{0.7670}  \\
                \bottomrule
            \end{tabular}
            }
            
\vspace{-2.2mm}
            \captionof{table}{\small Ablation study for different event filters. Case 3 is the baseline. } 
            \label{tab_abl2}

}
\hfill
\parbox{.34\textwidth}{
\centering

\scalebox{0.9}{
\begin{tabular}{c|c|cc}
                \toprule
                \multirow{2}{*}{\textbf{Case}}& \textbf{Backward}& \multirow{2}{*}{\textbf{PSNR}} & \multirow{2}{*}{\textbf{SSIM}} \\[0.5ex]
                ~  & \textbf{Injection}  & ~&~  \\
                \hline        
                6& - & 20.24&0.6920    \\
                7& w. Gating & 20.92&  0.7237 \\
                8& w. CA & 21.05&  0.7128\\
                \hline
                \textbf{Ours}  & w. BI & \textbf{23.63} & \textbf{0.7670}  \\
                \bottomrule
            \end{tabular}
            }
            
            \vspace{-2mm}
\captionof{table}{\small Ablation study for the design of Backward Injection (BI). Case 6 is the baseline. } 
\label{tab_abl3}
}
\vspace*{-1.5em}
\end{table*}

\vspace{-1mm}
\textbf{Comparison on Real-world Dataset.} 
As shown in \cref{tab_com_all} (SDE Dataset) and \cref{tab_com_RLE}, our method outperforms state-of-the-art (SOTA) techniques on real-world datasets. 
Specifically, the proposed method achieves higher performance in terms of PSNR and SSIM on both SDE-indoor/outdoor and RLE compared to image-based methods, demonstrating the benefits of incorporating paired event data for low-light enhancement. Compared to previous SOTA event-based methods EvLight \cite{Liang_2024_EvLight}, our method gets a large margin improvement of 0.89dB/1.01dB in terms of PSNR and the average improvement of 0.0286 in terms of SSIM on SDE-indoor/outdoor. On RLE, our approach further improves PSNR by 0.95 dB and SSIM by 0.0469. Meanwhile, our method requires only 9.4\% of the parameters used by EvLight, further demonstrating its efficiency. 
Furthermore, in \cref{fig:compare_rle} and \cref{fig:compare_sde}, we present the qualitative results of our method alongside other SOTA methods on RLE and SDE, respectively. Our method produces results with more coherent edges and natural colors while significantly reducing noise in the enhanced images.

\noindent \textbf{Comparison on Synthesized Dataset.} 
We conduct the same comparative experiments on the SDSD dataset. As shown in \cref{tab_com_all}, our method significantly outperforms image-based methods. Compared to EvLight, our approach also achieves enhancements of 1.24dB/0.0068 and 0.78dB/0.0312 in terms of PSNR/SSIM, respectively. As illustrated in \cref{fig:compare_sdsd}, the visual results demonstrate that our method restores more realistic and accurate detail textures. 

\subsection{Ablation Studies and Analysis}
\label{sec:exp_abl}
\textbf{Effects on Event and Illumination Guidance.} As demonstrated in \cref{tab_abl1}, 
the guidance of illumination (I) and events (E) are of significant importance for our method. While the elimination of these two components may result in a reduction of parameters and computations, the efficacy of the model is considerably diminished due to the absence of global illumination and HDR information. The incorporation of illumination and event guidance leads to improvements of 2.23dB/0.0383 and 2.55dB/0.0476 in PSNR/SSIM, respectively.

\noindent \textbf{Effects on IAEF.} We employ the IAEF module to selectively utilize informative events while mitigating the impact of noisy events in low-light scenarios. \cref{tab_abl2} demonstrate the necessity of this approach and the effectiveness of IAEF. Compared to the model without IAEF (Case 3), our method achieves a significant improvement of 2.71dB in PSNR. Simply using convolution (Case 4) or transformer block (Case 5) fails to suppress the influence of noise, as they neither leverage global illumination priors nor analyze the spatiotemporal differences between valid and noisy events. As shown in \cref{fig:vis_tsne}(d), IAEF effectively enhances noise suppression on event features. 

\noindent \textbf{Effects on Backward Injection.} The results in \cref{tab_abl3} indicate that Backward Injection is a reasonable and effective method. We remove it (Case 6) or replace it with gating (Case 7), as well as more complex mechanisms like cross-attention (Case 8), but these alternatives failed to extract the complementary information required for illumination and event features from the fused features. This may be due to the significant domain gap between the fused features and the modal-specific features without the assistance of the reused attention map $\mathbf{A}$. By employing Backward Injection, we achieved a PSNR improvement of over 2.58dB.

\section{Conclusion}
In this paper, we propose EIC-LIE, an event-illumination collaborative low-light image enhancement framework that addresses key limitations in existing event-based methods. Our Event-Illumination Collaborative Interaction (EICI) module enables the bidirectional interaction of event and illumination features, while the Illumination-Aware Event Filter (IAEF) suppresses event noise using brightness statistics. Furthermore, we construct a hybrid imaging system to collect high-quality event-image pairs, introducing the first high-resolution real-world event-based LIE dataset. Extensive experiments on real-world and synthetic datasets demonstrate that EIC-LIE surpasses state-of-the-art methods, validating its effectiveness in real-world low-light scenarios.

\section{Acknowledgement}
This work was supported by the National Natural Science Foundation of China (NSFC) under Grants 62225207, 62436008, 62422609 and 62276243. 




{
    \small
    \bibliographystyle{ieeenat_fullname}
    \bibliography{main}
}


\end{document}